\documentclass[onecolumn]{IEEEtran}
\IEEEoverridecommandlockouts
\usepackage{cite}
\usepackage{amsmath,amssymb,amsfonts}
\usepackage{algorithmic}
\usepackage{graphicx}
\usepackage{textcomp}
\usepackage{xcolor}
\usepackage{amsmath}
\usepackage{array}
\usepackage{booktabs}
\usepackage{float}
\usepackage{multirow}
\usepackage{makecell}
\usepackage{subcaption}
\usepackage{caption}
\usepackage{afterpage} 
\usepackage{caption}
\usepackage[numbers]{natbib}

\def\BibTeX{{\rm B\kern-.05em{\sc i\kern-.025em b}\kern-.08em
		T\kern-.1667em\lower.7ex\hbox{E}\kern-.125emX}}

\begin{document}

	\title{Cluster-Enhanced Federated Graph Neural Network for Recommendation\\
		\thanks{This research is supported by the National Natural Science Foundation of China under grant 62372385 (Corresponding Author: Y. Yuan).}
	}
	
\author{Haiyan Wang and Ye Yuan, \textit{Member, IEEE}}
	
	\maketitle
	
	\noindent\textbf{Abstract:} Personal interaction data can be effectively modeled as individual graphs for each user in recommender systems. Graph Neural Networks (GNNs)-based recommendation techniques have become extremely popular since they can capture high-order collaborative signals between users and items by aggregating the individual graph into a global interactive graph. However, this centralized approach inherently poses a threat to user privacy and security. Recently, federated GNN-based recommendation techniques have emerged as a promising solution to mitigate privacy concerns. Nevertheless, current implementations either limit on-device training to an unaccompanied individual graphs or necessitate reliance on an extra third-party server to touch other individual graphs, which also increases the risk of privacy leakage. To address this challenge, we propose a Cluster-enhanced Federated Graph Neural Network framework for Recommendation, named CFedGR, which introduces high-order collaborative signals to augment individual graphs in a privacy preserving manner. Specifically, the server clusters the pretrained user representations to identify high-order collaborative signals. In addition, two efficient strategies are devised to reduce communication between devices and the server. Extensive experiments on three benchmark datasets validate the effectiveness of our proposed methods.
	

	\section{Introduction}
	In recommendation systems, the interaction data for each user can be effectively represented as a individual graph structures \cite{Xu2023HRSTLR,alamgir2022federated,Bi2023TwoStreamGCN}. These personalized graphs capture the unique relationships and interactions between the user and various items, allowing for a more nuanced and accurate modeling of user preferences\cite{zhang2019deep,Luo2021ADMM,Yuan2024NodeCollaborationGCN}. The advanced capability of Graph Neural Networks (GNNs) to aggregate individual user interaction graphs into a comprehensive global interactive graph has led to their widespread application in recommender systems \cite{ying2018graph,Li2023MomentumLatentFactor,Wu2023DoubleSpaceQoS}. However, the centralized learning methodology necessitates the aggregation and centralized storage of substantial volumes of users' confidential data\cite{chang2021sequential,Jin2022NeuralDynamics,Yuan2022KalmanFilterLFA}. Despite its advantages, this method has several limitations: 1)The data involved is highly sensitive and the transmission of users' private interaction data beyond local devices poses a significant risk of privacy leakage\cite{Xia2015StochasticClouds,zhang2023comprehensive}. 2)Centralized data storage necessitates that third-party devices possess sufficient storage capacity\cite{Wang2024GraphTensorAttention}. 3)Stringent data protection regulations, such as the General Data Protection Regulation (GDPR)\cite{voigt2017eu,Yuan2024ProportionalIntegralLFA} have been enacted, prohibiting commercial entities from collecting, exchanging, or selling user data without explicit user consent. These limitations make it difficult to deploy centralized learning methods in real-life scenarios.
	
	Fortunately, the Federated Learning (FL) framework proposed by Google\cite{mcmahan2017communication,Yuan2024FuzzyPIDLFA} addresses these issues. FL allows multiple users to collaboratively train a global model, with users needing to upload only model parameters instead of raw data. In such scenarios, users are confined to constructing first-order subgraphs from their own interaction data, precluding the exploitation of information from other users with similar interests, specifically higher-order collaborative signals\cite{Li2023ManipulatorCalibration}. This constraint prevents GNNs from capturing high-order information, leading to degraded recommendation performance\cite{hu2014neighbors,Luo2021FastNMF,Wu2022DoubleSpaceQoS}. To address this issue, algorithms like FedGNN\cite{wu2022federated,Jiang2024RoleNegotiation} have been proposed, but it brought about risks of privacy leakage and increased communication burdens. Acquiring advanced collaborative signals while maintaining user privacy has emerged as a critical challenge.
	
	In light of the aforementioned challenges, we propose a Cluster-enhanced Federated Graph Neural Network framework for Recommendation(CFedGR). Within this framework, user features extracted by GNNs are aggregated into clusters. Given that these features integrate information from interacted items, users within the same cluster demonstrate analogous preferences, effectively serving as each other's collaborative context. To alleviate the communication overhead between server and local devices during the model training, we devise a sampling strategy to select representative users from each cluster for training, and restrict the number of collaborative neighbors to the top k most similar ones. To sum up, our contributions are summarized as follows:
	\begin{enumerate}
		\item We propose a cluster-enhanced federated graph neural network framework for recommendation, which effectively captures users' high-order collaborative information without compromising user privacy.
		\item We propose a clustering technique to augment users' local first-order interaction subgraphs and devise two straightforward yet effective strategies to alleviate communication overhead between server and devices.
		\item Extensive experiments, conducted on three benchmark datasets used in recommendation systems, have validated the effectiveness of our proposed methods.
	\end{enumerate}
	
	\section{RELATED WORKS}
	\subsection{Graph Neural Network}
	GNN have garnered extensive research attention in recent years owing to their exceptional capability to efficiently model graph data, which has prompted their wide adoption in different graph-based tasks\cite{wu2020comprehensive,9647958}, such as node classification\cite{rong2019dropedge,Luo2020PTPSO,Yuan2024AdaptiveDivergenceLFA}, link prediction\cite{rossi2021knowledge,Yang2024DataDrivenVibration}, and graph classification\cite{hong2020graph,Luo2022NeuLFT,Chen2024SDGNN}. By employing the technique of message passing defined as follows:\begin{equation}
		\mathbf{h}_v^{(k+1)} = \sigma \left( \mathbf{W} \cdot \text{AGGREGATE} \left( \{ \mathbf{h}_u^{(k)} : u \in \mathcal{N}(v) \} \right) \right)
	\end{equation} 
	where $\mathbf{h}_v^{(k+1)}$ represents the feature vector of node $v$ at layer $k + 1$, and $\mathbf{h}_u^{(k)}$ is the feature vector of node $u$ at layer $k$. The set $\mathcal{N}(v)$ consists of the neighbor nodes of node $v$. The \text{AGGREGATE} function collects information from the neighbors, utilizing methods such as summation, averaging, or taking the maximum. $\mathbf{W}$ is a matrix of learnable weights and \(\sigma\) signifies the activation function. GNN can iteratively combine the features of neighboring nodes with its own features to update the node's representation\cite{gao2023survey,Jin2023KWinnerTakeAll,Yuan2023KalmanLFA}. For example, Kipf et al.\cite{kipf2017semi} proposed a scalable approach for semi-supervised learning on graph-structured data using Graph Convolutional Networks (GCNs). It employs a localized first-order approximation of spectral graph convolutions to efficiently capture local graph structure and node features. Hamilton et al.\cite{hamilton2017inductive} introduced GraphSAGE, which enhances GNN by sampling and aggregating features from a node's local neighborhood. Veličković et al. \cite{velivckovic2017graph} presented Graph Attention Networks (GAT) that employ an attention mechanism to assign different importances to nodes in a neighborhood, allowing the model to focus on the most relevant parts of the graph. He et al.\cite{he2020lightgcn} proposed LightGCN, by eliminating feature transformation and nonlinear activation, they simplify traditional GCNs, creating a more efficient model that focuses solely on neighborhood aggregation.
	\subsection{Federated Recommendation Systems}
	Federated Learning (FL) allows multiple clients to collaboratively train machine learning models while safeguarding data privacy and has seen a significant rise in application to recommendation systems in recent years, which is known as Federated Recommender Systems (FedRecs) \cite{he2021fedgraphnn,Yuan2022KalmanFilterLFA,Yuan2023AdaptiveDivergenceModel}. In this paradigm, rather than transmitting their private data, clients engage in the training of local model using data stored locally and subsequently transmit their model parameters or gradients to a central server, which then aggregates this information to derive a comprehensive global model,  effectively mitigating the risk of data leakage\cite{Li2023RobotArmCalibration}. FedRecs can generally be categorized into two approaches: federated recommendation based on matrix factorization and federated recommendation based on graph neural networks. Matrix factorization is a technique commonly used in recommendation systems. It works by decomposing the user-item rating matrix into two lower-dimensional matrices\cite{Luo2021LearningDepthQoS,Yuan2022MultilayeredLatentFactor}, the user matrix U and the item matrix V, such that their product approximates the original rating matrix. For example, Federated Collaborative Filtering (FCF) \cite{chai2020secure} suggested each client utilizes local data and item embeddings retrieved from the server to update user embeddings, ensuring that user data never leaves the client device. Du,Yongjie et al.\cite{du2021federated} introduced a randomized response mechanism to further enhance privacy protection. Recommendation methods based on graph neural networks leverage these networks to derive latent representations of users and items, which are consequently employed for downstream recommendation tasks. For instance, FedSoG\cite{liu2022federated} adopts a relational attention mechanism to distinguish the contributions of neighboring users and interacted items to the user's embedding, effectively addressing heterogeneity in the process. A similar work to ours is FedGNN \cite{wu2022federated}. It utilizes a trusted third-party server and homomorphic encryption to identify users' neighbors and augment the local subgraph. However, this approach still carries the risk of leaking user privacy and requires higher communication overhead.
	\section{PROPOSED METHOD}
	\begin{figure*}[htbp]
		\centering
		\includegraphics[width=\textwidth]{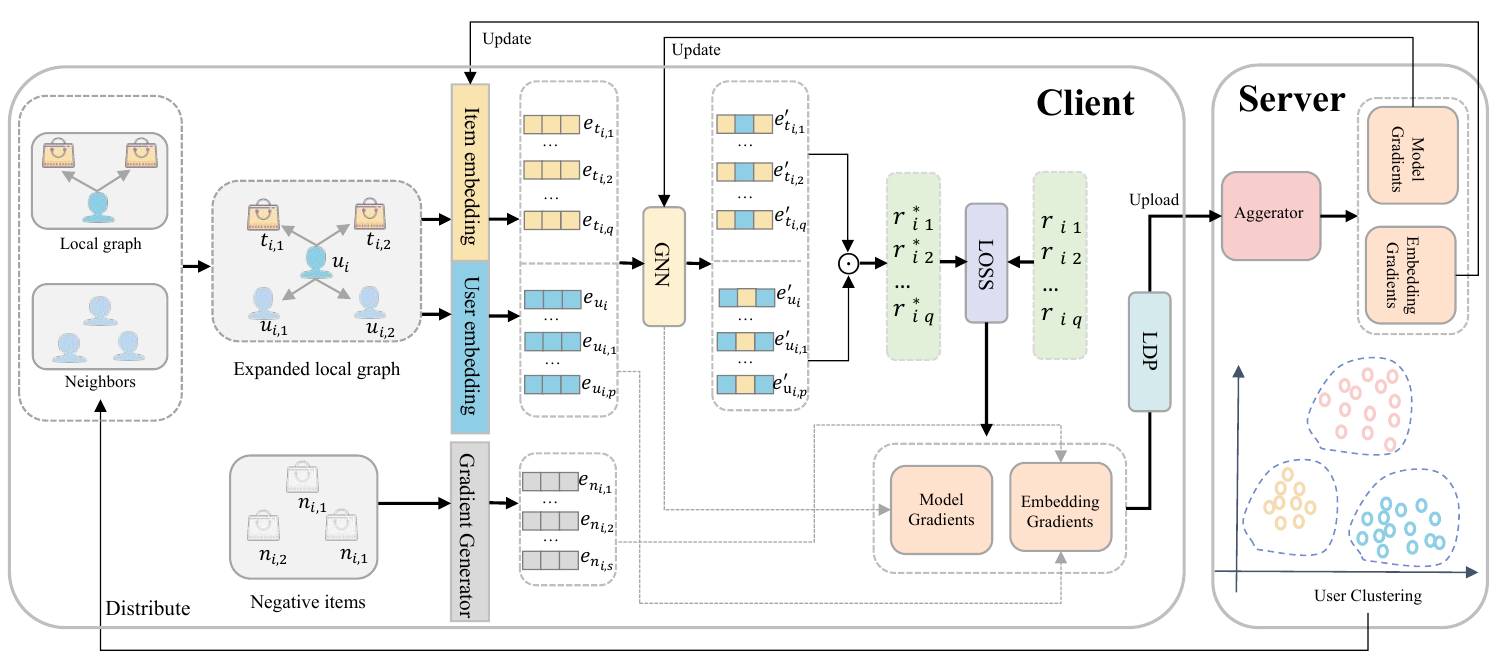} 
		\caption{Framework of the CFedGR}
		\label{framework}
	\end{figure*}
	\subsection{Problem Formulation}
	Consider \( U = \{ u_1, u_2, \ldots, u_M \} \) and \( T = \{ t_1, t_2, \ldots, t_N \} \) as the sets of users and items, with \( M \) and \( N \) representing the number of users and items, respectively. The embeddings of users and items are denoted by \(\mathbf{E}_u \in \mathbb{R}^{M \times d}\) and \(\mathbf{E}_i \in \mathbb{R}^{N \times d}\), where \( d \) indicates the dimensionality of these vectors. Denote the neighbors of user \( u_i \) as \( N_i = \{ n_{i,1}, n_{i,2}, \ldots, n_{i,p} \} \), where \( p \) represents the number of neighboring users. Users assign ratings to the items they interact with. For example, the rating given by user \(i\) to item \(j\) is denoted as \(r_{ij}\). For each user \( u_i \), let \( t_i = \{ t_{i,1}, t_{i,2}, \ldots, t_{i,q} \} \) represent the set of items with which the user has interacted, with the corresponding ratings represented as \(r_i= \{ r_{i1}, r_{i2}, \ldots, r_{iq} \} \), where \( q \) denotes the number of items the user has rated. Based on the interaction data from all users, a rating matrix \( \mathbf{R} \in \mathbb{R}^{M \times N} \)can be obtained. To mitigate the risk of data leakage, users' interaction data is retained locally on their devices. Accordingly, there exists solely a first-order user-item graph, consisting solely of each user and the items they have interacted with. Our objective is to identify each user's collaborative neighboring users without compromising user privacy and utilize the neighbors' information to predict the ratings(\( r \in \mathbf{R}\setminus \sum_{i \in U} r_i \)) for items that the users have not previously interacted with .
	\subsection{Overview of the Framework}
	This section introduces the proposed framework, as illustrated in the Fig. \ref{framework}. The framework consists of two main components: the client-side and the server-side. The client-side is responsible for training a local GNN with data stored on local devices. We represent the embeddings of the user, the user's neighbors, the items interacted with and the negative items by the user as \( e_{u_i} \), \( \{ e_{u_{i,1}}, e_{u_{i,2}}, \ldots, e_{u_{i,p}} \} \),  \( \{ e_{t_{i,1}}, e_{t_{i,2}}, \ldots, e_{t_{i,q}} \} \), and\( \{ e_{n_{i,1}}, e_{n_{i,2}}, \ldots, e_{n_{i,s}} \} \), respectively. Subsequently clients upload both the embedding gradients and model gradients to the server. The server is tasked with three main responsibilities: determining the users eligible for the training, aggregating the gradients uploaded by the clients to update global model and embeddings, and discerning the collaborative neighbors associated with each user. Next, we will detail the coordination mechanism between the server-side and client-side.
	
	At the outset of each training round, the server selects a subset of users for participation. An embedding layer is deployed on the server to map user IDs and item IDs into embeddings. Acknowledging the initial imprecision of user embeddings, T rounds of pre-training are performed without utilizing neighbor embeddings. Following this phase, the server undertakes clustering based on the user embeddings to discern the neighbors for each user. Upon obtaining the embeddings of users, collaborative neighbors, and interacted items, the client feeds these embeddings into the local GNN to derive more expressive embeddings. In the subsequent step, the predicted ratings are computed by taking the inner product of the user embeddings  and item embeddings according to \eqref{rating}.
	\begin{equation}
		\hat{r}_{ij} = \mathbf{u}_i \cdot \mathbf{v}_j\label{rating}
	\end{equation}
	After obtaining all the predicted ratings \((\hat{r}_{i1}, \hat{r}_{i2}, \ldots, \hat{r}_{iq})\) for user \( u_i \), we use  \eqref{difference} to compute the loss with the original ratings \( (r_{i1}, r_{i2}, \ldots, r_{iq}) \).
	\begin{equation}
		\text{Loss} = \frac{1}{q} \sum_{k=1}^{q} (r_{ik} - \hat{r}_{ik})^2\label{difference}
	\end{equation}
	\(\theta\) is defined as the GNN parameters. Derived from the loss function, the gradients of GNN, user embeddings and item embeddings \(\nabla_{\theta} L\), \(\nabla_{E_u} L\), and \(\nabla_{E_i} L\) are then uploaded to the server. Once the gradients are received from all clients, the server aggregates them using the following formula:
	\begin{equation}
		\mathbf{\nabla \mathcal{L}}' = \sum_{k=1}^{N} \frac{n_k}{n} \nabla \mathcal{L}_k
	\end{equation}
	where \( n_k \) is the number of interaction data for user \( k \), and \( n \) is the total number of interaction data  for all users, the \(\nabla \mathcal{L}_k\) is the gradients of user \(u_k\).
	We designate the learning rate as \( l \). According to \eqref{model}, \eqref{user}, and \eqref{item}, the updated GNN parameters \(\theta'\), user embeddings \(\mathbf{E}_u'\) and item embeddings \(\mathbf{E}_i'\) are obtained. 
	\begin{equation}
		\theta' = \theta - l \nabla_{\theta} \mathcal{L}'\label{model}
	\end{equation}
	\begin{equation}
		\mathbf{E}_u' = \mathbf{E}_u - l \nabla_{\mathbf{E}_u} \mathcal{L}'\label{user}
	\end{equation}
	\begin{equation}
		\mathbf{E}_i' = \mathbf{E}_i - l \nabla_{\mathbf{E}_i} \mathcal{L}'\label{item}
	\end{equation}
	Finally the server broadcasts the updated GNN and embeddings to the clients for the next round of training. In the following sections, we provide a detailed explanation of several important modules within the framework.
	\subsection{Local User-Item Graph Augmentation via Clustering}
	In the context of federated learning, each user exclusively retains data pertaining to the items they have interacted with, acquiring information about a user's collaborative neighbors while ensuring their privacy remains intact is a considerable challenge. Under current methodologies such as FedGNN \cite{wu2022federated}, which incorporate an additional third-party server alongside the central server and utilize homomorphic encryption, clients are able to upload encrypted user embeddings and item IDs. However, this approach not only escalates the deployment costs and network communication burden but also poses a significant privacy risk if the central and third-party servers collude. To address this issue, we propose a clustering-based method to expand the local user-item graph, which can reveal user's neighbors while preserving user privacy.
	
	Firstly, the server that holds all user embeddings performs clustering on these embeddings. To prevent performance degradation caused by inaccurate user embeddings at the initial stage, we conduct clustering after pre-training for \(T\) rounds. Numerous clustering algorithms available for this purpose, including the K-means algorithm\cite{hartigan1979algorithm,Hu2023FCAN}, hierarchical clustering\cite{ward1963hierarchical,Hu2021LinkClustering,Hu2021LinkClustering}, and the DBSCAN algorithm\cite{ester1996density,Hu2020ClusterAlgorithm}. In our method, we employ the K-means clustering algorithm. After that, users with similar interests are grouped into the same cluster. Except for the user themselves, the remaining users in the cluster are considered the user's collaborative neighbors. Proceeding with this, the IDs of other users in the cluster are sent to each user by server, and edges are added between the user and the received neighbor nodes. Throughout this process, the server does not receive any private data from the users, nor is there any data sharing between users, which ensures that the local first-order user-item graph can be expanded while maintaining user privacy.
	\subsection{Strategies for Communication Efficiency}
	To further reduce the communication overhead between the server and devices, we design two simple yet effective methods. Firstly, instead of the random user selection commonly used in most federated recommendation systems, we cluster all user embeddings on the server-side and select a proportional number of users from each cluster to participate in the training. This ensures that the model is exposed to a wider variety of user types and behavior patterns during training, which aids in better learning and generalization of the model. Secondly, during the user neighbor matching phase, the server does not return the IDs of all users within the same cluster, as not all of them contribute to constructing the user's preferences. Instead, server selects the top k users with the highest similarity to the user based on the similarity matrix, which helps to further reduce the communication between the server and the users.
	\subsection{User Privacy Protection}
	In the model training phase, users need to upload gradients for interacted items. However, only items with which the user has interacted have non-zero item gradients. As a result, the server can potentially reconstruct the user's interaction history based on these non-zero gradients. To safeguard against the disclosure of users' interaction histories, we sample \(n \) negative items from those not interacted with by the user and generate gradients for these negative samples using a Gaussian distribution. These generated gradients are then uploaded along with the actual item gradients, making it difficult for the server to distinguish the user's true interaction history. Denoted the real item gradients as \(\mathbf{g}_{\text{real}}\), with the mean and variance of these gradients represented by \(\mathbb{E}[\mathbf{g}_{\text{real}}]\) and \(\mathbb{V}[\mathbf{g}_{\text{real}}]\), respectively.  \(\mathbf{z}\) is a random vector following a normal distribution. According to \eqref{gussion},we can obtain the gradients for the negative samples.
	\begin{equation}
		\mathbf{g}_{\text{neg}} = \mathbb{E}[\mathbf{g}_{\text{real}}] + \sqrt{\mathbb{V}[\mathbf{g}_{\text{real}}]} \cdot \mathbf{z}\label{gussion}
	\end{equation}
	
	As the model gradients uploaded by users contain private information such as user preferences, they also pose a risk of privacy leakage. To mitigate this, we employ local differential privacy by adding noise to these gradients before uploading them. This ensures that the server cannot infer users' private information from the gradients.
	\begin{equation}
		g' = \text{clip}(g, \delta) + \text{Laplace}(0, \lambda)\label{ldp}
	\end{equation}
	where \(\text{clip}(g, \delta)\) clips the gradient \(g\) to ensure its norm does not exceed \(\delta\), \(\text{Laplace}(0, \lambda)\) is Laplace noise with mean 0 and scale \(\lambda\).
	\section{EXPERIMENTS}
	\renewcommand{\arraystretch}{1.2}
	\subsection{Experimental Settings and Comparisons}
	In our experiments,  three benchmark datasets in recommendation systems are selected to evaluate the recommendation performance: FilmTrust, MovieLens-100k and Douban. The statistical information of these datasets is shown in Table \ref{datasets}.
\begin{table}[H]
	\caption{Statistics of the Dataset}
	\begin{center}
		\setlength{\tabcolsep}{10pt} 
		\renewcommand{\arraystretch}{1.5} 
		\begin{tabular}{|c|c|c|c|c|}
			\hline
			\textbf{Dataset} & \textbf{Users} & \textbf{Items} & \textbf{Ratings} & \textbf{Sparsity} \\ \hline
			Filmtrust & 1508 & 2071 & 35497 & 98.85\% \\ \hline
			MovieLens-100K & 943 & 1682 & 100000 & 93.70\% \\ \hline
			Douban & 3000 & 3000 & 136891 & 98.47\% \\ \hline
		\end{tabular}
		\label{datasets}
	\end{center}
\end{table}

	Each dataset is split into training, validation, and test sets with a ratio of \(70\%\), \(10\%\), and \(20\%\), respectively. The learning rate is optimized within the range of \{0.01, 0.05, 0.1\}, while the decay rate is explored within \{0.0001, 0.0005, 0.001\}. Each iteration engages 256 users, with embeddings mapped into 128 dimensions. We establish a gradient clipping threshold of 0.2 and set the noise scale to 0.1. Additionally, the \( k \)-value is set to 200 and 1000 negative samples are utilized. A two-layer Graph Attention Network (GAT) serves as the local GNN. Performance is assessed using two classic metrics: Mean Absolute Error (MAE)\cite{Luo2021ADMM} and Root Mean Square Error (RMSE)\cite{Wu2020QoSPrediction,Li2022MomentumLFA}, with lower values indicating better results. Baseline models from both centralized and federated learning frameworks are employed to evaluate recommendation efficacy.

	\noindent\textbf{Centralized Learning:}
	\begin{itemize}
		\item \textbf{NCF}\cite{He2017Neural}: It combines matrix factorization and multi-layer perceptron to learn the interaction function between users and items.
		\item \textbf{GraphRec}\cite{Rashed2019}: It integrates graph feature information into matrix factorization.
		\item \textbf{Glocal-K}\cite{Han2021GLocalKGA}: This approach pre-trains a localized kernelized weight matrix and trains an autoencoder.Then fine-tunes the autoencoder using global convolutional kernels.
	\end{itemize}

	\begin{table}[htbp]
		\caption{Performance comparison of different methods}
		\begin{center}
			\renewcommand{\arraystretch}{1.5} 
			\setlength{\tabcolsep}{10pt} 
			\begin{tabular}{|c|cc|cc|cc|}
				\hline
				\multirow{2}{*}[0.5ex]{\centering \textbf{Method}} & \multicolumn{2}{c|}{\textbf{Filmtrust}} & \multicolumn{2}{c|}{\textbf{MovieLens-100K}} & \multicolumn{2}{c|}{\textbf{Douban}} \\
				\cline{2-7} 
				& \textbf{MAE} & \textbf{RMSE} & \textbf{MAE} & \textbf{RMSE} & \textbf{MAE} & \textbf{RMSE} \\
				\hline
				NCF & 0.6351 & 0.8174 & 0.7354 & 0.9321 & 0.5964 & 0.7540 \\
				GraphRec & 0.6554 & 0.8561 & 0.7128 & 0.9033 & \textbf{0.5788} & \textbf{0.7379} \\
				GLocal-K & \textbf{0.6146} & \textbf{0.8026} & \textbf{0.6933} & \textbf{0.8895} & 0.5826 & 0.7428 \\
				\hline
				MetaMF & 0.6901 & 0.8756 & 0.7941 & 0.9990 & 0.6537 & 0.8259 \\
				FedNCF & 0.6854 & 0.8613 & 0.7890 & 0.9922 & 0.6470 & 0.8134 \\
				FedGNN & \textbf{0.6313} & \textbf{0.8162} & \textbf{0.7705} & \textbf{0.9646} & \textbf{0.6477} & \textbf{0.8036} \\
				Ours & 0.6351 & 0.8170 & 0.7725 & 0.9661 & 0.6484 & 0.8067 \\
				\hline
			\end{tabular}
		\end{center}
		\label{performance_comparison}
	\end{table}

	\begin{figure}[htbp]
		\centering
		\begin{subfigure}[b]{0.35\textwidth}
			\includegraphics[width=\textwidth]{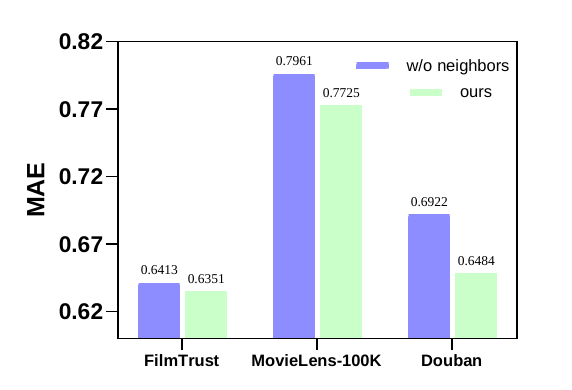}
			\caption{MAE performance}
			\label{fig:epinions}
		\end{subfigure}
		\hspace{1pt}
		\begin{subfigure}[b]{0.35\textwidth}
			\includegraphics[width=\textwidth]{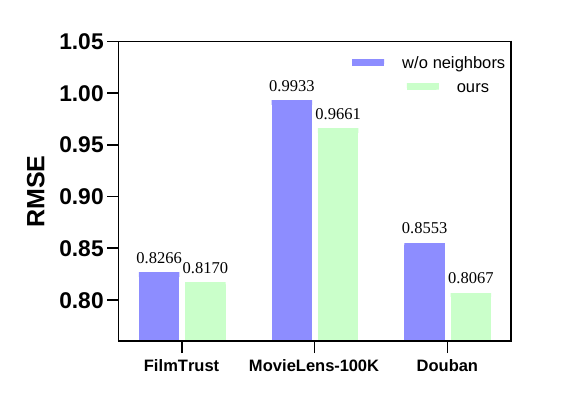}
			\caption{RMSE performance}
			\label{fig:ciao}
		\end{subfigure}
		\caption{Ablation study without collaborative neighbors}
		\label{ablation}
	\end{figure}

	\clearpage
	\textbf{Federated Learning}:
	\begin{itemize}
		\item \textbf{FedNCF}\cite{Perifanis2022Federated}: It is a federated version of NCF, where user and item embeddings are shared on the server.
		\item \textbf{MetaMF}\cite{lin2020meta}: It incorporates a collaborative memory module and a meta-recommendation module for producing personalized item embeddings and recommendation models.
		\item \textbf{FedGNN}\cite{wu2022federated}: It utilizes a third-party server and homomorphic encryption to find neighboring users, which enhances the feature representation of users.
	\end{itemize}
	\begin{table}[tpb]
		\caption{Performance comparison on different values of N}
		\begin{center}
			\renewcommand{\arraystretch}{1.5} 
			\setlength{\tabcolsep}{10pt} 
			\begin{tabular}{|c|cc|cc|}
				\hline
				\multirow{2}{*}{\textbf{of clusters}} & \multicolumn{2}{c|}{\textbf{MovieLens-100K}} & \multicolumn{2}{c|}{\textbf{Douban}} \\
				\cline{2-5}
				& \textbf{MAE} & \textbf{RMSE} & \textbf{MAE} & \textbf{RMSE} \\
				\hline
				N=5 & 0.7822 & 0.9771 & 0.6846 & 0.8476 \\
				\hline
				N=10 & 0.7826 & 0.9779 & 0.6839 & 0.8456 \\
				\hline
				N=20 & 0.7815 & 0.9767 & 0.6831 & 0.8445 \\
				\hline
			\end{tabular}
			\label{clusters_comparison}
		\end{center}
	\end{table}
	\subsection{Analysis of Experimental Results}
	The experimental results on all datasets are shown in Table \ref{performance_comparison}, from which we can make the following observations:
		\begin{table}[htpb]
		\caption{Number of neighbors required by FedGNN
		}
		\centering
		\begin{tabular}{|c|c|}
			\hline
			\textbf{Datasets} & \textbf{of neighbors} \\
			\hline
			Filmtrust & 1094 \\
			\hline
			Douban & 815 \\
			\hline
			MovieLens-100K & 860 \\
			\hline
		\end{tabular}
		\label{neighbors}
	\end{table}
	
		\begin{figure}[tpb]
		\centering
		\begin{subfigure}[b]{0.35\linewidth}
			\includegraphics[width=\linewidth]{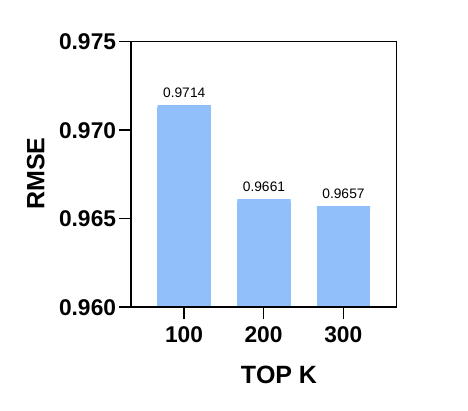}
			\caption{MovieLens-100K}
			\label{fig:epinions}
		\end{subfigure}
		\hspace{0.02\linewidth} 
		\begin{subfigure}[b]{0.35\linewidth}
			\includegraphics[width=\linewidth]{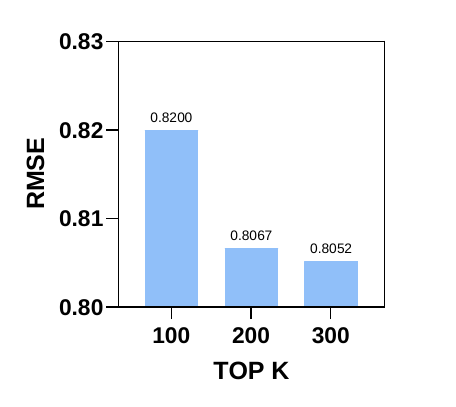}
			\caption{Douban}
			\label{fig:ciao}
		\end{subfigure}
		\caption{Performance comparison on different values of K}
		\label{two_graphs}
	\end{figure}
	
	\begin{enumerate}
		\item Although centralized learning methods achieve the lowest error rates, they pose significant risks to user privacy and face considerable challenges in real-life deployment due to data collection difficulties. In contrast, our method only slightly compromises performance while ensuring user privacy, thus offering a superior trade-off between privacy protection and recommendation accuracy.
		
		\item In contrast to other federated recommendation models that do not incorporate high-order collaborative neighbor information, such as MetaMF and FedNCF, our model delivers superior prediction performance across all datasets. This demonstrates the effectiveness of our proposed framework and highlights the critical role of neighbor information in enhancing recommendation systems.
		\item Despite FedGNN showing marginally better than our model, the difference is minimal, with a maximum improvement of only 0.3\%. However, FedGNN necessitates an additional server, which not only heightens the risk of privacy leakage but also increases server deployment costs and communication burdens. Furthermore, as shown in Table \ref{neighbors}, the number of neighbors that FedGNN needs to transmit is significantly higher than in our model. Our method reduces neighbor communication by approximately 4 to 5 times, achieving a superior balance between communication overhead and recommendation performance.
	\end{enumerate}
	\subsection{Hyperparameter Analysis}

	In this section, we discuss the impacts of hyperparameters, specifically the number of clusters \( N \) and the value of top \( K \).
	We conducted experiments on two relatively large datasets, MovieLens-100K and Douban. Table \ref{clusters_comparison} demonstrates that performance remains stable across various cluster numbers when the top \( K \) is fixed at 40, which alleviates the burden of parameter tuning.
	Furthermore, as illustrated in Fig. \ref{two_graphs}, RMSE gradually decreases with increasing \( K \). However, once \( K \) reaches 300, the reduction in error becomes negligible. To minimize communication overhead, we recommend setting the \( K \)-value to 200.
	\subsection{Ablation Study}
	To verify the effectiveness of the clustering algorithm, we conducted experiments under identical settings without its use. The results, presented in Fig. \ref{ablation}, show that performance is significantly worse without the clustering algorithm. This demonstrates the effectiveness of our clustering algorithm and underscores the importance of high-order collaborative neighbor information in recommendation systems.
	\section{Conclusion}
	In this paper, we emphasize the significance of high-order collaborative information in federated recommendation systems and propose a cluster-enhanced federated graph neural network framework for secure item recommecdation. Our proposed framework clusters users with similar preferences together to derive high-order collaborative neighbors while preserving user privacy. To alleviate communication burden, we designed a sampling strategy that selects representative users from each cluster for each training round. Additionally, the server captures only the top k neighbors with the highest similarity to the user within the cluster, instead of considering all neighbors. Comprehensive experiments on three datasets confirm that our framework achieves an exemplary balance between privacy protection and recommendation performance.

	\bibliography{refe}

\begin{thebibliography}{10}
\providecommand{\url}[1]{#1}
\csname url@samestyle\endcsname
\providecommand{\newblock}{\relax}
\providecommand{\bibinfo}[2]{#2}
\providecommand{\BIBentrySTDinterwordspacing}{\spaceskip=0pt\relax}
\providecommand{\BIBentryALTinterwordstretchfactor}{4}
\providecommand{\BIBentryALTinterwordspacing}{\spaceskip=\fontdimen2\font plus
\BIBentryALTinterwordstretchfactor\fontdimen3\font minus
  \fontdimen4\font\relax}
\providecommand{\BIBforeignlanguage}[2]{{%
\expandafter\ifx\csname l@#1\endcsname\relax
\typeout{** WARNING: IEEEtran.bst: No hyphenation pattern has been}%
\typeout{** loaded for the language `#1'. Using the pattern for}%
\typeout{** the default language instead.}%
\else
\language=\csname l@#1\endcsname
\fi
#2}}
\providecommand{\BIBdecl}{\relax}
\BIBdecl

\bibitem{Xu2023HRSTLR}
X.~Xu, M.~Lin, X.~Luo, and Z.~Xu, ``{HRST-LR: A Hessian regularization
  spatio-temporal low rank algorithm for traffic data imputation},'' \emph{IEEE
  Transactions on Intelligent Transportation Systems}, vol.~24, no.~10, pp.
  11\,001--11\,017, 2023.

\bibitem{alamgir2022federated}
Z.~Alamgir, F.~K. Khan, and S.~Karim, ``Federated recommenders: methods,
  challenges and future,'' \emph{Cluster Computing}, vol.~25, no.~6, pp.
  4075--4096, 2022.

\bibitem{Bi2023TwoStreamGCN}
F.~Bi, T.~He, Y.~Xie, and X.~Luo, ``{Two-stream graph convolutional
  network-incorporated latent feature analysis},'' \emph{IEEE Transactions on
  Services Computing}, vol.~16, no.~4, pp. 3027--3042, 2023.

\bibitem{zhang2019deep}
S.~Zhang, L.~Yao, A.~Sun, and Y.~Tay, ``Deep learning based recommender system:
  A survey and new perspectives,'' \emph{ACM computing surveys (CSUR)},
  vol.~52, no.~1, pp. 1--38, 2019.

\bibitem{Luo2021ADMM}
X.~Luo, Y.~Zhong, Z.~Wang, and M.~Li, ``{An alternating-direction-method of
  multipliers-incorporated approach to symmetric non-negative latent factor
  analysis},'' \emph{IEEE Transactions on Neural Networks and Learning
  Systems}, vol.~34, no.~8, pp. 4826--4840, 2021.

\bibitem{Yuan2024NodeCollaborationGCN}
Y.~Yuan, Y.~Wang, and X.~Luo, ``{A Node-Collaboration-Informed Graph
  Convolutional Network for Highly Accurate Representation to Undirected
  Weighted Graph},'' \emph{IEEE Transactions on Neural Networks and Learning
  Systems}, 2024.

\bibitem{ying2018graph}
R.~Ying, R.~He, K.~Chen, P.~Eksombatchai, W.~L. Hamilton, and J.~Leskovec,
  ``Graph convolutional neural networks for web-scale recommender systems,'' in
  \emph{Proceedings of the 24th ACM SIGKDD international conference on
  knowledge discovery \& data mining}, 2018, pp. 974--983.

\bibitem{Li2023MomentumLatentFactor}
W.~Li, X.~Luo, H.~Yuan, and M.~Zhou, ``{A Momentum-accelerated
  Hessian-vector-based Latent Factor Analysis Model},'' \emph{IEEE Transactions
  on Services Computing}, vol.~16, no.~2, pp. 830--844, 2023.

\bibitem{Wu2023DoubleSpaceQoS}
D.~Wu, P.~Zhang, Y.~He, and X.~Luo, ``{A Double-Space and Double-Norm Ensembled
  Latent Factor Model for Highly Accurate Web Service QoS Prediction},''
  \emph{IEEE Transactions on Services Computing}, vol.~16, no.~2, pp. 802--814,
  2023.

\bibitem{chang2021sequential}
J.~Chang, C.~Gao, Y.~Zheng, Y.~Hui, Y.~Niu, Y.~Song, D.~Jin, and Y.~Li,
  ``Sequential recommendation with graph neural networks,'' in
  \emph{Proceedings of the 44th international ACM SIGIR conference on research
  and development in information retrieval}, 2021, pp. 378--387.

\bibitem{Jin2022NeuralDynamics}
L.~Jin, X.~Zheng, and X.~Luo, ``{Neural dynamics for distributed collaborative
  control of manipulators with time delays},'' \emph{IEEE/CAA Journal of
  Automatica Sinica}, vol.~9, no.~5, pp. 854--863, 2022.

\bibitem{Yuan2022KalmanFilterLFA}
Y.~Yuan, X.~Luo, M.~Shang, and Z.~Wang, ``{A Kalman-filter-incorporated latent
  factor analysis model for temporally dynamic sparse data},'' \emph{IEEE
  Transactions on Cybernetics}, vol.~53, no.~9, pp. 5788--5801, 2022.

\bibitem{Xia2015StochasticClouds}
Y.~N. Xia, M.~C. Zhou, X.~Luo, S.~C. Pang, and Q.~S. Zhu, ``{Stochastic
  modeling and performance analysis of migration-enabled and error-prone
  clouds},'' \emph{IEEE Transactions on Industrial Informatics}, vol.~11,
  no.~2, pp. 495--504, 2015.

\bibitem{zhang2023comprehensive}
S.~Zhang, W.~Yuan, and H.~Yin, ``Comprehensive privacy analysis on federated
  recommender system against attribute inference attacks,'' \emph{IEEE
  Transactions on Knowledge and Data Engineering}, 2023.

\bibitem{Wang2024GraphTensorAttention}
L.~Wang, K.~Liu, and Y.~Yuan, ``{GT-A2T: Graph Tensor Alliance Attention
  Network},'' \emph{IEEE/CAA Journal of Automatica Sinica}, 2024.

\bibitem{voigt2017eu}
P.~Voigt and A.~Von~dem Bussche, ``The eu general data protection regulation
  (gdpr),'' \emph{A Practical Guide, 1st Ed., Cham: Springer International
  Publishing}, vol.~10, no. 3152676, pp. 10--5555, 2017.

\bibitem{Yuan2024ProportionalIntegralLFA}
Y.~Yuan, S.~Lu, and X.~Luo, ``{A Proportional Integral Controller-Enhanced
  Non-negative Latent Factor Analysis Model},'' \emph{IEEE/CAA Journal of
  Automatica Sinica}, 2024.

\bibitem{mcmahan2017communication}
B.~McMahan, E.~Moore, D.~Ramage, S.~Hampson, and B.~A. y~Arcas,
  ``Communication-efficient learning of deep networks from decentralized
  data,'' in \emph{Artificial intelligence and statistics}.\hskip 1em plus
  0.5em minus 0.4em\relax PMLR, 2017, pp. 1273--1282.

\bibitem{Yuan2024FuzzyPIDLFA}
Y.~Yuan, J.~Li, and X.~Luo, ``{A Fuzzy PID-Incorporated Stochastic Gradient
  Descent Algorithm for Fast and Accurate Latent Factor Analysis},'' \emph{IEEE
  Transactions on Fuzzy Systems}, vol.~32, no.~7, pp. 4049--4061, 2024.

\bibitem{Li2023ManipulatorCalibration}
Z.~Li, S.~Li, O.~Bamasag, A.~Alhothali, and X.~Luo, ``{Diversified
  Regularization Enhanced Training for Effective Manipulator Calibration},''
  \emph{IEEE Transactions on Neural Networks and Learning Systems}, vol.~34,
  no.~11, pp. 8778--8790, 2023.

\bibitem{hu2014neighbors}
L.~Hu, A.~Sun, and Y.~Liu, ``Your neighbors affect your ratings: on
  geographical neighborhood influence to rating prediction,'' in
  \emph{Proceedings of the 37th international ACM SIGIR conference on Research
  \& development in information retrieval}.\hskip 1em plus 0.5em minus
  0.4em\relax ACM, 2014, pp. 345--354.

\bibitem{Luo2021FastNMF}
X.~Luo, Y.~Zhou, Z.~Liu, and M.~C. Zhou, ``{Fast and accurate non-negative
  latent factor analysis of high-dimensional and sparse matrices in recommender
  systems},'' \emph{IEEE Transactions on Knowledge and Data Engineering},
  vol.~35, no.~4, pp. 3897--3911, 2021.

\bibitem{Wu2022DoubleSpaceQoS}
D.~Wu, P.~Zhang, Y.~He, and X.~Luo, ``{A double-space and double-norm ensembled
  latent factor model for highly accurate web service QoS prediction},''
  \emph{IEEE Transactions on Services Computing}, vol.~16, no.~2, pp. 802--814,
  2022.

\bibitem{wu2022federated}
C.~Wu, F.~Wu, L.~Lyu, T.~Qi, Y.~Huang, and X.~Xie, ``A federated graph neural
  network framework for privacy-preserving personalization,'' \emph{Nature
  Communications}, vol.~13, no.~1, p. 3091, 2022.

\bibitem{Jiang2024RoleNegotiation}
Q.~Jiang, D.~Liu, H.~Zhu, S.~Wu, N.~Wu, X.~Luo, and Y.~Qiao, ``{Iterative Role
  Negotiation via the Bi-level GRA++ with Decision Tolerance},'' \emph{IEEE
  Transactions on Computational Social Systems}, 2024.

\bibitem{wu2020comprehensive}
Z.~Wu, S.~Pan, F.~Chen, G.~Long, C.~Zhang, and S.~Y. Philip, ``A comprehensive
  survey on graph neural networks,'' \emph{IEEE transactions on neural networks
  and learning systems}, vol.~32, no.~1, pp. 4--24, 2020.

\bibitem{9647958}
X.~Luo, H.~Wu, Z.~Wang, J.~Wang, and D.~Meng, ``A novel approach to large-scale
  dynamically weighted directed network representation,'' \emph{IEEE
  Transactions on Pattern Analysis and Machine Intelligence}, vol.~44, no.~12,
  pp. 9756--9773, 2022.

\bibitem{rong2019dropedge}
Y.~Rong, W.~Huang, T.~Xu, and J.~Huang, ``Dropedge: Towards deep graph
  convolutional networks on node classification,'' \emph{arXiv preprint
  arXiv:1907.10903}, 2019.

\bibitem{Luo2020PTPSO}
X.~Luo, Y.~Yuan, S.~Chen, N.~Zeng, and Z.~Wang, ``{Position-transitional
  particle swarm optimization-incorporated latent factor analysis},''
  \emph{IEEE Transactions on Knowledge and Data Engineering}, vol.~34, no.~8,
  pp. 3958--3970, 2020.

\bibitem{Yuan2024AdaptiveDivergenceLFA}
Y.~Yuan, X.~Luo, and M.~Zhou, ``{Adaptive Divergence-based Non-negative Latent
  Factor Analysis of High-Dimensional and Incomplete Matrices from Industrial
  Applications},'' \emph{IEEE Transactions on Emerging Topics in Computational
  Intelligence}, vol.~8, no.~2, pp. 1209--1222, 2024.

\bibitem{rossi2021knowledge}
A.~Rossi, D.~Barbosa, D.~Firmani, A.~Matinata, and P.~Merialdo, ``Knowledge
  graph embedding for link prediction: A comparative analysis,'' \emph{ACM
  Transactions on Knowledge Discovery from Data (TKDD)}, vol.~15, no.~2, pp.
  1--49, 2021.

\bibitem{Yang2024DataDrivenVibration}
W.~Yang, S.~Li, and X.~Luo, ``{Data Driven Vibration Control: A Review},''
  \emph{IEEE/CAA Journal of Automatica Sinica}, 2024.

\bibitem{hong2020graph}
D.~Hong, L.~Gao, J.~Yao, B.~Zhang, A.~Plaza, and J.~Chanussot, ``Graph
  convolutional networks for hyperspectral image classification,'' \emph{IEEE
  Transactions on Geoscience and Remote Sensing}, vol.~59, no.~7, pp.
  5966--5978, 2020.

\bibitem{Luo2022NeuLFT}
X.~Luo, H.~Wu, and Z.~Li, ``{NeuLFT: A novel approach to nonlinear canonical
  polyadic decomposition on high-dimensional incomplete tensors},'' \emph{IEEE
  Transactions on Knowledge and Data Engineering}, vol.~35, no.~6, pp.
  6148--6166, 2022.

\bibitem{Chen2024SDGNN}
J.~Chen, Y.~Yuan, and X.~Luo, ``{SDGNN: Symmetry-Preserving Dual-Stream Graph
  Neural Networks},'' \emph{IEEE/CAA Journal of Automatica Sinica}, vol.~11,
  no.~7, pp. 1717--1719, 2024.

\bibitem{gao2023survey}
C.~Gao, Y.~Zheng, N.~Li, Y.~Li, Y.~Qin, J.~Piao, Y.~Quan, J.~Chang, D.~Jin,
  X.~He \emph{et~al.}, ``A survey of graph neural networks for recommender
  systems: Challenges, methods, and directions,'' \emph{ACM Transactions on
  Recommender Systems}, vol.~1, no.~1, pp. 1--51, 2023.

\bibitem{Jin2023KWinnerTakeAll}
L.~Jin, S.~Liang, X.~Luo, and M.~Zhou, ``{Distributed and Time-Delayed
  k-Winner-Take-All Network for Competitive Coordination of Multiple Robots},''
  \emph{IEEE Transactions on Cybernetics}, vol.~53, no.~1, pp. 641--652, 2023.

\bibitem{Yuan2023KalmanLFA}
Y.~Yuan, X.~Luo, M.~Shang, and Z.~Wang, ``{A Kalman-Filter-Incorporated Latent
  Factor Analysis Model for Temporally Dynamic Sparse Data},'' \emph{IEEE
  Transactions on Cybernetics}, vol.~53, no.~9, pp. 5788--5801, 2023.

\bibitem{kipf2017semi}
T.~N. Kipf and M.~Welling, ``Semi-supervised classification with graph
  convolutional networks,'' in \emph{International Conference on Learning
  Representations (ICLR)}, 2017.

\bibitem{hamilton2017inductive}
W.~Hamilton, Z.~Ying, and J.~Leskovec, ``Inductive representation learning on
  large graphs,'' \emph{Advances in neural information processing systems},
  vol.~30, 2017.

\bibitem{velivckovic2017graph}
P.~Veli{\v{c}}kovi{\'c}, G.~Cucurull, A.~Casanova, A.~Romero, P.~Lio, and
  Y.~Bengio, ``Graph attention networks,'' \emph{arXiv preprint
  arXiv:1710.10903}, 2017.

\bibitem{he2020lightgcn}
X.~He, K.~Deng, X.~Wang, Y.~Li, Y.~Zhang, and M.~Wang, ``Lightgcn: Simplifying
  and powering graph convolution network for recommendation,'' in
  \emph{Proceedings of the 43rd International ACM SIGIR conference on research
  and development in Information Retrieval}, 2020, pp. 639--648.

\bibitem{he2021fedgraphnn}
C.~He, K.~Balasubramanian, E.~Ceyani, C.~Yang, H.~Xie, L.~Sun, L.~He, L.~Yang,
  S.~Y. Philip, Y.~Rong \emph{et~al.}, ``Fedgraphnn: A federated learning
  benchmark system for graph neural networks,'' in \emph{ICLR 2021 Workshop on
  Distributed and Private Machine Learning (DPML)}, 2021.

\bibitem{Yuan2023AdaptiveDivergenceModel}
Y.~Yuan, R.~Wang, G.~Yuan, and X.~Luo, ``{An Adaptive Divergence-based
  Non-negative Latent Factor Model},'' \emph{IEEE Transactions on Systems, Man,
  and Cybernetics: Systems}, vol.~53, no.~10, pp. 6475--6487, 2023.

\bibitem{Li2023RobotArmCalibration}
Z.~Li, S.~Li, and X.~Luo, ``{A Novel Machine Learning System for Industrial
  Robot Arm Calibration},'' \emph{IEEE Transactions on Circuits and Systems II:
  Express Briefs}, 2023.

\bibitem{Luo2021LearningDepthQoS}
X.~Luo, M.~Chen, H.~Wu, Z.~Liu, H.~Yuan, and M.~C. Zhou, ``{Adjusting learning
  depth in nonnegative latent factorization of tensors for accurately modeling
  temporal patterns in dynamic QoS data},'' \emph{IEEE Transactions on
  Automation Science and Engineering}, vol.~18, no.~4, pp. 2142--2155, 2021.

\bibitem{Yuan2022MultilayeredLatentFactor}
Y.~Yuan, Q.~He, X.~Luo, and M.~Shang, ``{A Multilayered-and-Randomized Latent
  Factor Model for High-Dimensional and Sparse Matrices},'' \emph{IEEE
  Transactions on Big Data}, vol.~8, no.~3, pp. 784--794, 2022.

\bibitem{chai2020secure}
D.~Chai, L.~Wang, K.~Chen, and Q.~Yang, ``Secure federated matrix
  factorization,'' \emph{IEEE Intelligent Systems}, vol.~36, no.~5, pp. 11--20,
  2020.

\bibitem{du2021federated}
Y.~Du, D.~Zhou, Y.~Xie, J.~Shi, and M.~Gong, ``Federated matrix factorization
  for privacy-preserving recommender systems,'' \emph{Applied soft computing},
  vol. 111, p. 107700, 2021.

\bibitem{liu2022federated}
Z.~Liu, L.~Yang, Z.~Fan, H.~Peng, and P.~S. Yu, ``Federated social
  recommendation with graph neural network,'' \emph{ACM Transactions on
  Intelligent Systems and Technology (TIST)}, vol.~13, no.~4, pp. 1--24, 2022.

\bibitem{hartigan1979algorithm}
J.~A. Hartigan and M.~A. Wong, ``Algorithm as 136: A k-means clustering
  algorithm,'' \emph{Journal of the royal statistical society. series c
  (applied statistics)}, vol.~28, no.~1, pp. 100--108, 1979.

\bibitem{Hu2023FCAN}
L.~Hu, Y.~Yang, Z.~Tang, Y.~He, and X.~Luo, ``{FCAN-MOPSO: An improved
  fuzzy-based graph clustering algorithm for complex networks with
  multiobjective particle swarm optimization},'' \emph{IEEE Transactions on
  Fuzzy Systems}, vol.~31, no.~10, pp. 3470--3484, 2023.

\bibitem{ward1963hierarchical}
J.~H. Ward~Jr, ``Hierarchical grouping to optimize an objective function,''
  \emph{Journal of the American statistical association}, vol.~58, no. 301, pp.
  236--244, 1963.

\bibitem{Hu2021LinkClustering}
L.~Hu, J.~Zhang, X.~Pan, X.~Luo, and H.~Yuan, ``{An effective link-based
  clustering algorithm for detecting overlapping protein complexes in
  protein-protein interaction networks},'' \emph{IEEE Transactions on Network
  Science and Engineering}, vol.~8, no.~4, pp. 3275--3289, 2021.

\bibitem{ester1996density}
M.~Ester, H.-P. Kriegel, J.~Sander, X.~Xu \emph{et~al.}, ``A density-based
  algorithm for discovering clusters in large spatial databases with noise,''
  in \emph{kdd}, vol.~96, no.~34, 1996, pp. 226--231.

\bibitem{Hu2020ClusterAlgorithm}
L.~Hu, S.~Yang, X.~Luo, and M.~C. Zhou, ``{An algorithm of inductively
  identifying clusters from attributed graphs},'' \emph{IEEE Transactions on
  Big Data}, vol.~8, no.~2, pp. 523--534, 2020.

\bibitem{Wu2020QoSPrediction}
D.~Wu, X.~Luo, M.~Shang, Y.~He, G.~Wang, and X.~Wu, ``{A
  data-characteristic-aware latent factor model for web services QoS
  prediction},'' \emph{IEEE Transactions on Knowledge and Data Engineering},
  vol.~34, no.~6, pp. 2525--2538, 2020.

\bibitem{Li2022MomentumLFA}
W.~Li, X.~Luo, H.~Yuan, and M.~C. Zhou, ``{A momentum-accelerated
  Hessian-vector-based latent factor analysis model},'' \emph{IEEE Transactions
  on Services Computing}, vol.~16, no.~2, pp. 830--844, 2022.

\bibitem{He2017Neural}
X.~He, L.~Liao, H.~Zhang, L.~Nie, X.~Hu, and T.-S. Chua, ``Neural collaborative
  filtering.'' in \emph{The Web Conference}, 2017.

\bibitem{Rashed2019}
A.~Rashed, J.~Grabocka, and L.~Schmidt-Thieme, ``Attribute-aware non-linear
  co-embeddings of graph features,'' in \emph{Conference on Recommender
  Systems}, 2019.

\bibitem{Han2021GLocalKGA}
S.~C. Han, T.~Lim, S.~Long, B.~Burgstaller, and J.~Poon, ``Glocal-k: Global and
  local kernels for recommender systems,'' \emph{Proceedings of the 30th ACM
  International Conference on Information \& Knowledge Management}, 2021.

\bibitem{Perifanis2022Federated}
V.~Perifanis and P.~S. Efraimidis, ``Federated neural collaborative
  filtering,'' \emph{Knowledge-based systems}, vol. 242, p. 108441, 2022.

\bibitem{lin2020meta}
Y.~Lin, P.~Ren, Z.~Chen, Z.~Ren, D.~Yu, J.~Ma, M.~d. Rijke, and X.~Cheng,
  ``Meta matrix factorization for federated rating predictions,'' in
  \emph{Proceedings of the 43rd International ACM SIGIR Conference on Research
  and Development in Information Retrieval}, 2020, pp. 981--990.

\end{thebibliography}

\end{document}